\documentclass[a4paper,review,11pt,authoryear]{elsarticle}
\usepackage{amsmath,bm,hyperref,url,fullpage,mathtools,booktabs,bbm}
\usepackage{paralist,enumitem,todonotes,amssymb}
\usepackage{multirow}
\usepackage{threeparttable}
\usepackage{graphicx}
\usepackage[ruled,linesnumbered]{algorithm2e}
\usepackage{breakurl}
\usepackage{caption}
\usepackage{subcaption}
\usepackage{rotating}
\usepackage[a4paper, margin=1in]{geometry}
\usepackage{algorithm2e}
\usepackage{hyperref}
\hypersetup{
	colorlinks=true,
	linkcolor=blue,
	citecolor=blue,
	urlcolor=blue}
\newcommand{\x}{\boldsymbol{x}}
\newcommand{\D}{\mathcal{D}}

\newcommand{\expect}{e_{\tau}}

\newcommand{\hattheta}{\widehat{\boldsymbol{\theta}}}
\newcommand{\haty}{\widehat{y}_i}

\newcommand{\checkf}{\rho_{\tau}}
\newcommand{\E}{\mathbb{E}}
\newcommand{\R}{\mathbb{R}}
\newcommand{\sumN}{\sum_{i=1}^n}
\newcommand{\sumH}{\sum_{h=1}^H}

\begin{document}
	\date{}
	\begin{frontmatter}	
\title{Neural Networks for Censored Expectile Regression Based on Data Augmentation}
		\author[1]{Wei Cao} 
		\author[1]{Shanshan Wang}
        \ead{Corresponding author at (School of Economics and Management, Beihang University, Beijing 100191, China) via sswang@buaa.edu.cn.}
		
		\address[1]{School of Economics and Management, Beihang
			University, Beijing, China}
		
		\begin{abstract}
Expectile regression neural networks (ERNNs) are powerful tools for capturing heterogeneity and complex nonlinear structures in data. However, most existing research has primarily focused on fully observed data, with limited attention paid to scenarios involving censored observations. In this paper, we propose a data augmentation–based ERNNs algorithm, termed \textsf{DAERNN}, for modeling heterogeneous censored data. The proposed \textsf{DAERNN} is fully data-driven, requires minimal assumptions, and offers substantial flexibility. Simulation studies and real-data applications demonstrate that \textsf{DAERNN} outperforms existing censored ERNNs methods and achieves predictive performance comparable to models trained on fully observed data. Moreover, the algorithm provides a unified framework for handling various censoring mechanisms without requiring explicit parametric model specification, thereby enhancing its applicability to practical censored data analysis.
		\end{abstract}

		
	\end{frontmatter}
	
\section{Introduction}
\label{sec:introduction}

Expectile regression (ER), first proposed by \citet{newey_asymmetric_1987}, has been extensively studied as a flexible alternative to quantile regression (QR) for modeling heterogeneous distributions. By estimating expectiles across different levels, one can obtain an alternative characterization of the conditional distribution—similar to QR, but with smoother behavior. 
Specifically, for $\tau\in (0,1)$, the $\tau$th regression expectile of $Y$ given $X$ is obtained by:
\begin{align}\label{eq:er}
	\expect(Y|\x)&=\mathop{\arg\min}_\theta \E\Big([\checkf(Y-\theta)-\checkf(Y)]|X=\x\Big),
\end{align}
where $\checkf(u)=\frac{1}{2}\cdot|\tau-\textbf{1}(u<0)|\cdot u^2$ is the expectile check function, where $\textbf{1}(\cdot)$ is the indicator function. 
Unlike QR, ER uses a differentiable squared loss function rather than the non-differentiable absolute loss, resulting in greater computational efficiency. ER also simplifies inference because its asymptotic distribution can be derived without estimating the error density function \citep{newey_asymmetric_1987, abdous1995relating}. Moreover, the asymmetric squared loss enables ER to capture both the probability and magnitude of tail behaviors, providing valuable insights into the distribution’s shape. This makes ER particularly effective for quantifying tail-related risks and estimating conditional expectiles at specified probability levels, supporting high-impact risk assessment and expectation-based evaluation. For example, \citet{taylor2008estimating} and \citet{kuan_assessing_2009} use expectiles to estimate Value-at-Risk (VaR) and to construct conditional autoregressive models for assessing extreme losses. Recent work has extended expectile-based methods to a range of applications, including risk measures \citep{chavez2016extreme, kim2016nonlinear, daouia_estimation_2018, mohammedi2021consistency, xu_prediction_2022}, exchange rate volatility \citep{xie2014varying}, stock index and portfolio modeling \citep{sahamkhadam2021dynamic, jiang2022single}, and high-frequency financial data analysis \citep{gerlach2022bayesian}. \citet{seipp_weighted_2021}, \citet{ciuperca2025right}, and \citet{zhang2024weighted} also apply ER to survival analysis, particularly for estimating mean residual life expectancy. Clearly, there is rising interest in ER. 

With the rapid advancement of information technology, data acquisition has become increasingly efficient, but the relationships among variables have grown more complex. In many cases, linear models fail to capture these nonlinear structures. Machine learning methods have therefore gained significant attention for their flexibility and ability to model complex variable interactions without strong parametric assumptions. Building on this, numerous studies have extended ER with machine learning techniques to better capture nonlinear patterns. For example, \citet{yang2015nonparametric} proposed a tree-based gradient boosting method for nonparametric multiple ER, while \citet{jiang_expectile_2017} introduced the Expectile Regression Neural Network (ERNNs). Similarly, \citet{farooq2017svm} and \citet{pei2021novel} developed support vector machine–based ER models (ER-SVM), and \citet{cai2023expectile} integrated random forests with ER to create the ERF model. More recently, \citet{yang2024nonparametric} extended ERNNs to high-dimensional settings, enabling nonlinear variable selection. These approaches demonstrate strong performance in modeling complex, nonlinear data structures.

However, the above studies primarily focus on fully observed datasets. In real-world applications, data incompleteness is common, with censoring being one of the most frequent cases. Censoring occurs when the event of interest is not observed within the study period. Such data are common in fields like biomedicine \citep{wu_cure_2013, wu_multiple_2017, yu_quantile_2021, narisetty_censored_2022}, engineering management \citep{chen2020predictive, xu2023lifespan, xiao2024improving}, and economics \citep{Nico_unemployment, kiefer1988economic, ludemann_censored_2006}. Censoring introduces discrepancies between observed and true population values, complicating the modeling of conditional distributions. To address this challenge, various statistical methods have been developed, including the Kaplan–Meier (KM) estimator \citep{kaplan1958nonparametric}, the Cox proportional hazards model \citep{cox_regression_1972}, and the accelerated failure time (AFT) model \citep{wei_accelerated_1992}, along with numerous extensions. More recently, machine learning techniques have been incorporated into survival analysis to capture complex nonlinear relationships, such as neural networks–based Cox models \citep{faraggi1995neural}, hazard models \citep{zhong2021deep}, AFT models \citep{kim2024deep}, and censored quantile regression models \citep{jia2022deep, hao2023data}.

Despite growing interest in survival analysis with deep learning, expectile regression neural networks (ERNNs) for censored data remain largely unexplored. \citet{zhang2024weighted} proposed a neural networks–based approach for right-censored data using inverse probability weighting (IPW), termed \textsf{WERNN} (Weighted Expectile Regression Neural Networks). However, \textsf{WERNN} is limited to right-censoring, whereas many real-world applications involve other censoring types, such as left \citep{wang2018tobit} or interval censoring \citep{choi2024interval}. Moreover, the IPW approach requires estimating the censoring distribution, which is computationally intensive in high dimensions and can be unreliable when the survival function is miss-specified.

To our knowledge, apart from \citet{zhang2024weighted}, no other studies have addressed censored ERNNs. Furthermore, the existing method can only accommodate a single censoring type. This highlights the need for a unified estimation framework for ERNNs that can flexibly handle multiple censoring mechanisms.
Inspired by the data augmentation strategy of \citet{yang2018new}, we propose a novel algorithm, the \textbf{D}ata \textbf{A}ugmentation \textbf{E}xpectile \textbf{R}egression \textbf{N}eural \textbf{N}etworks (\textsf{DAERNN}). Starting from an initial model, \textsf{DAERNN} iteratively updates through three key steps—data augmentation, model updating, and prediction—and typically converges within a few iterations. Notably, the method is highly flexible and can be applied to various censoring mechanisms, including singly left-, right-, and interval-censored data.

The remainder of this paper is structured as follows. In Section~\ref{sec:method} we introduce the proposed method \textsf{DAERNN}. Section \ref{sec:experiments} conducts Monte Carlo simulations to investigate the finite sample performance of the proposed \textsf{DAERNN} method. Real case applications are shown in Section~\ref{sec:realdata}. Section~\ref{sec:discussion} gives a brief conclusion of this article and also provides some potential research directions about ERNNs for censored data.

\section{Methodology}\label{sec:method}
In this section, we introduce the proposed \textsf{DAERNN} method, which is designed to effectively handle censored datasets exhibiting complex nonlinear relationships and heteroscedastic structures. 
\subsection{Model setup}\label{sub:2-1}

Assume that $\{y_i,\x_i\}_{i=1}^n$ are samples draw from $(Y,X)$, where $\x_i=(1,x_{i1},\cdots,x_{ip})^T$ includes the $p$-dimensional covariates. To better capture the intricate relationship of the conditional expectile between independent and dependent variables, we assume the $\tau$th conditional expectile of $Y$ given $X$ defined in Eq.\eqref{eq:er} takes the form:
\begin{align}\label{eq:nlner}
	\expect(y_i|\x_i)&= m_\tau(\x_i), \ i=1,\cdots,n,
\end{align}
where $m_\tau:\R^p \rightarrow \R$ is an unknown smoothing function. Compare with traditional linear model, model~\eqref{eq:nlner} imposes no parametric assumptions on the functional form, thereby offering greater flexibility in capturing complex and nonlinear structures inherent in the data. However, in practical applications involving censored data, the estimation of model~\eqref{eq:nlner} faces two major challenges. Fisrt, although $m_\tau(\cdot)$ enhances flexibility for modeling complex data, the nonparametric nature simultaneously increases the difficulty of estimation. In addition, in the presence of censoring, the response variable $y_i$ is only partially observed, resulting in biased samples that cannot be directly used in standard expectile regression. 

Due to censoring, the observed response $t_i$ may not coincide with the true latent response $y_i$. 
Let $\delta_i$ denote the censoring type for the $i$th response, which is defined as 1) $\delta_i=0$: no censoring; 2) $\delta_i=1$: right censoring at $R_i$; 3) $\delta_i=2$: left censoring at $L_i$; and 4) $\delta_i=3$: interval censoring between $(L_i,R_i)$. Here $L_i$ and $R_i$ denote the left and right censoring points, respectively. Thus, 
the observed response $t_i$ can be expressed as
\begin{align*}
	t_i&=\left\{\begin{array}{ll}
		y_i,  &  \text{no censoring}\\
		y_i \wedge R_i,  &  \text{right censoring at}\ R_i\\
		y_i \vee L_i,  &   \text{left censoring at}\ L_i\\
		L_i \vee (y_i\wedge R_i),  &  \text{interval censoring between}\ (L_i,R_i)
	\end{array}\right.,
\end{align*}
where 
\(\wedge\) and \(\vee\) denote the minimum and maximum operations. Consequently, model estimation can only be based on the available observations $\{t_i, \x_i,\delta_i\}$ for $ i = 1, 2, \ldots, n$. To overcome the challenges discussed above, we develop a novel algorithm that integrates a neural networks framework with data augmentation techniques and more details are discussed in Section~\ref{sub:2-2}.

\subsection{Data-Augmented Expectile Regression Neural Networks for Censored Data:  DAERNN}\label{sub:2-2}

In this section, we first introduce the \textsf{DAERNN} procedure for estimating $m_\tau(\cdot)$ in (\ref{eq:nlner}) under censoring. The procedure consists of three main steps: data augmentation, model updating, and prediction. We then describe the estimation process for the model updating step, specifically focusing on the ERNNs model.

\subsubsection{General framework of \textsf{DAERNN}}\label{model}
Estimating $m_\tau(\cdot)$ under censoring is challenging because it requires recovering an unknown nonlinear function while correcting for bias introduced by censoring. Traditional nonparametric methods, such as kernel regression, local polynomials, and smoothing splines, can model nonlinear structures but struggle with highly complex relationships and suffer from the well-known “curse of dimensionality” \citep{hardle1994applied}. 
Neural networks offer a flexible alternative, capable of capturing complex nonlinear patterns in high-dimensional settings. As composite functions of affine transformations and nonlinear activation maps, organized into multiple layers, neural networks can approximate highly complex functions \citep{hornik1989multilayer,hornik1991approximation}. Deeper architectures further enhance their ability to represent intricate or high-dimensional structures \citep{telgarsky2016benefits,yarotsky2017error}. Building on this, we adopt the ERNNs model to estimate $m_\tau(\cdot)$, which will be described in Section \ref{sub:Estimation}. 

To address censoring, we integrate a data augmentation strategy that imputes censored observations, avoiding the need to explicitly estimate the survival function. This approach is naturally adaptable to various censoring mechanisms and offers both flexibility and computational efficiency. Prior studies have demonstrated that imputing event times can effectively reduce bias, particularly under complex censoring scenarios \citep{hsu2007multiple,lee2012multiple,yang2018new,hao_damcqrnn_2023,cao2024expectile}.

Algorithm~\ref{alg:alg1} summarizes the procedure of \textsf{DAERNN}. To implement the algorithm, two hyper-parameters must be specified: the imputation expectile levels $\{\tau_k = \frac{k}{m+1}: k =1,\cdots,m\}$ with $m$ being a prespecified integer, i.e, $m=\max\{\lfloor \sqrt{n}\rfloor,99\}$ in our proposal, and the number of iterations $H$. In addition, we define $S(i)$ as the potential set of real response $y_i$:
\begin{equation*}
	S(i)=\left\{
	\begin{array}{ll}
		{y_i}, & \text{no censoring}\\
		(R_i,\infty), & \text{right censoring at}\ R_i\\
		(-\infty,L_i) , & \text{left censoring at}\ L_i\\
		(L_i,R_i) , & \text{interval censoring between}\ (L_i,R_i)
	\end{array}
	\right.,
\end{equation*}
which is essential for the imputation step. As shown in Algorithm~\ref{alg:alg1}, \textsf{DAERNN} consists of two main stages: Initialization and Iterative Estimation.

\textsf{Initialization}. The goal of this step is to train an initial model that imputes censored outcomes and generates preliminary estimates for subsequent iterations. To improve computational efficiency, hyperparameter tuning is performed only on uncensored observations, and the selected parameters are then applied to train the ERNNs model for the remainder of the procedure. In this initialization step, we obtain $m$ ERNNs models, denoted as $M^{(0)}(\tau_k)$ for $\tau_k, \, k = 1, \ldots, m$.

\textsf{Iterative estimation at the $h$th Step}. Each iteration of \textsf{DAERNN} involves three sub-steps: data augmentation, model updating, and prediction.
\begin{enumerate}
    \item Data augmentation. For each censored sample, we impute a value from the predicted set $\{\hat{t}_i^{(h)} : i = 1, \ldots, n,\, \delta_i \neq 0\}$ obtained in the $(h-1)$-th iteration. The imputed $\tilde{t}_i^{(h)}$ value is accepted if it satisfies the censoring constraint $S(i)$; otherwise, the sampling is repeated until a feasible value is found. If no feasible value can be drawn, the censored sample is imputed using its censoring boundary, that is, $L_i, R_i, or (L_i+R_i)/2$ for left, right, and interval censoring, respectively.
    \item Model Updating. After imputation, the ERNNs models are retrained using the augmented dataset $\{\tilde{t}_i^{(h)},\x_i\}$, yielding updated models $M^{(h)}(\tau_k)$ for $k = 1, \ldots, m$. These models will be used for prediction in the next iteration.
    \item Prediction. The updated models $M^{(h)}(\tau_k)$ generate predictions for all expectile levels $\tau_k$ on the test data.
\end{enumerate}

\begin{algorithm}[ht]
    \small
	\caption{DAERNN: Data-Augmented Expectile Regression Neural Networks for Censored Data}
	\label{alg:alg1}
	\SetKwData{Left}{left}\SetKwData{This}{this}\SetKwData{Up}{up}
	\SetKwFunction{Union}{Union}\SetKwFunction{FindCompress}{FindCompress}
	\SetKwInOut{Input}{Input}\SetKwInOut{Output}{Output}
	
	\Input{Observed training data $\D_{train}=\{t_i, \x_i, \delta_i,S(i)\}$,  a grid of expectile levels $\{\tau_k, k=1,\cdots m\}$ , a prespecied number of iteration steps $H$ and predictors of testing set $\x_i^{test}$.}
	
	\Output{The $\tau_k$th conditional expectile predictor: $\haty(\tau_k)=\frac{1}{H}\sumH\haty^{(h)}(\tau_k)$}

        \textbf{Initialization}: 
	\emph{Build ERNNs models $M^{(0)}(\tau_k)$ for expectile level $\tau_k$ with un-censored observations $\{(t_i, \x_i) : i = 1, \ldots, n,\,\delta_i = 0 \}$}
	
	\For{$h=1,2,...,H$}{
		\textbf{Data Augmentation} : 
		\emph{Fitted the censored samples $\{\hat{t}_i^{(h)}(\tau_k): \delta_i \ne 0\}$ with $M^{(h-1)}(\tau_k)$ and define those expectiles fitted value as alternative sampling set.}
		
		\emph{Randomly draw a value from the alternative sampling set that satisfies the censoring constraint $S(i)$. Combined with the uncensored responses, these values form the pseudo fully observed responses $\tilde{t}_i^{(h)}$, defined as:
        $$\tilde{t}_i^{(h)} =\begin{cases}
        t_i, & \delta_i = 0 \\
        \hat t_i^{(h)} \in S(i), & \delta_i \neq 0~
        \end{cases}.$$}
		\textbf{Model Updating}:
		\emph{Update the ERNNs model with data after augmentation $\{\tilde{t}_i^{(h)},\boldsymbol{x}_i\}$ denoted as $M^{(h)}(\tau_k)$}
        
            \textbf{Prediction}: 
            \emph{Predict the responses $\haty^{(h)}(\tau_k)$ of testing set with the testing predictors through the updated ERNNs model $M^{(h)}(\tau_k)$}	
     }

\end{algorithm}

The above process is repeated until the predefined number of iterations $H$ is reached, and the final prediction is obtained by averaging the results across all iterations.

\subsubsection{Estimation of ERNNs model}\label{sub:Estimation}

Following the general framework above, we now present the estimation of $m_\tau(\cdot)$ under the proposed \textsf{DAERNN} structure. We implement a multilayer perceptron (MLP) structure, where the hidden-layer activations $g_{i,l}$ are computed as
\begin{equation}\label{eq:ERNN}
\begin{aligned}
    g_{i,1}&=f_1\left(\boldsymbol{x}_i^\top w_{i}^{(j,1)}+b^{(1)}\right),\\
    g_{i,l}&=f_l\left(\boldsymbol{x}_i^\top w_{i}^{(j,l)}+b^{(l)}\right), l=2,\cdots,L,
\end{aligned}
\end{equation}
where $w^{(l)}$ and $b^{(l)}$ are the weight and bias of layer $l$, respectively. The activation function $f_l$ is typically chosen as the Sigmoid or Rectified Linear Unit (ReLU). With data augmentation, the censored dataset is reconstructed as $\{\tilde{t}_i^{(h)}, \x_i\}$ which is then treated as pseudo fully observed data. Based on the neural network structure defined in Equation (\ref{eq:ERNN}), the optimal model parameters are obtained by solving the following optimization problem:
\begin{equation}\label{eq:ERNN-sol}
\begin{aligned}
&\hattheta=\mathop{\arg\min}\limits_{\theta}\frac{1}{n}\sumN\checkf\big(\tilde{t}_i^{(h)}-\hat{t}_i^{(h)}),\\
&\hat{t}_i^{(h)}=f_{o}\left(g_{i,L}^\top w_{i}^{\left(o\right)}+b^{\left(o\right)}\right),
\end{aligned}
\end{equation}
where $\hat{t}_i^{(h)}$ is the prediction value at $h$ step, $n$ is the total number of samples, $L$ is the number of hide layer, $w_{i}^{(o)}$ and $b^{(o)}$ represents the weights of and  bias term of output layer. The model parameter vector is given by $\hattheta=\big[w_{i}^{(o)},b^{(o)},w_{i}^{(j,l)},b^{(l)}\big]^\top$.

Unlike the QR, the expectile loss $\checkf(\cdot)$ is differentiable at all expectile levels, which facilitates efficient optimization using gradient-based methods. In this work, we employ mini-batch gradient descent (MBGD), which balances the efficiency of stochastic gradient descent with the stability of full-batch methods \citep{dekel2012optimal}. MBGD computes gradients on small random batches, improving computational efficiency, reducing gradient variance, and enabling parallel or distributed training for large-scale problems \citep{dekel2012optimal, khirirat2017mini}.

In practice, hyperparameter tuning is crucial for effective neural network modeling. While increasing the number of hidden layers and the number of nodes per layer can enhance the capacity of model to capture complex nonlinear relationships, it also raises the risk of overfitting. In this study, we consider the following hyperparameters: number of hidden layers, nodes per layer, dropout rate and training epochs. Additionally, we include the mini-batch size as a hyperparameter arising from the use of MBGD. We adopt a grid search with cross-validation to select the optimal combination, following prior studies \citep{jiang_expectile_2017, zhang2024weighted}.

\begin{figure}[ht]
    \centering
    \includegraphics[width=1\linewidth]{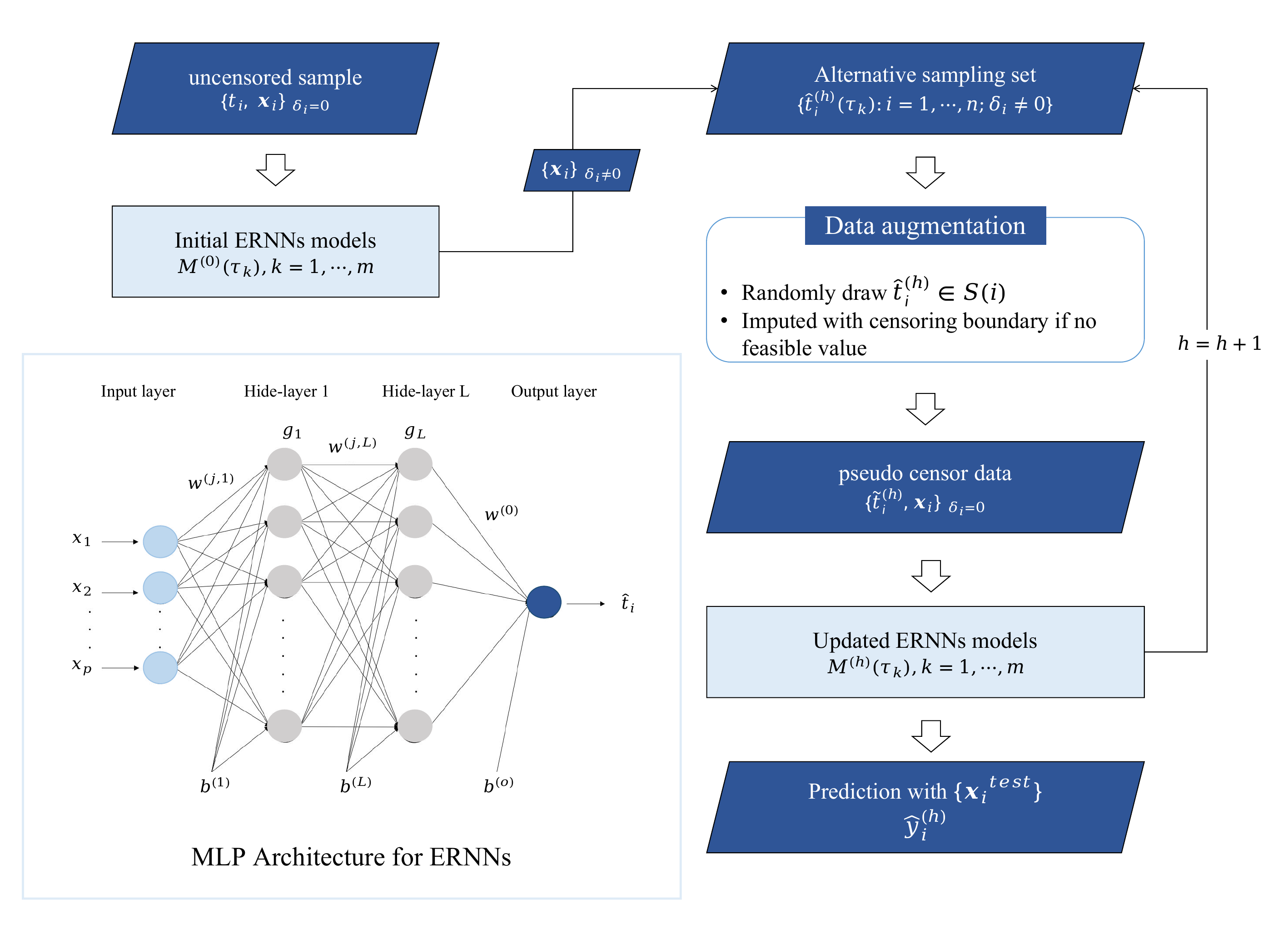}
    \caption{Flowchart of \textsf{DAERNN} Algorithm}
    \label{fig:DAERNN}
\end{figure}

In summary, we also illustrate the flowchart of the proposed algorithm in Figure \ref{fig:DAERNN}.

\section{Simulation Studies}
\label{sec:experiments}

In this section, we conduct simulation studies to evaluate the predictive performance of our proposed method. For comparison, we focus primarily on the \textsf{DAERNN} method and a neural networks model that does not account for censoring, i.e., it ignores the influence of censoring, referred to as the \textsf{FULL}. 
In addition, we compare against two existing methods for censored expectile regression: the \textsf{DALinear} method \citep{cao2024expectile} and the Weighted Expectile Regression Neural Networks (\textsf{WERNN}) method \citep{zhang2024weighted}.

\subsection{Data generation} \label{sub:3-1}

We consider the following homoscedastic and heteroscedastic cases, respectively:
\begin{itemize}
    \item []\textbf{(1) Model 1 (Homoscedastic Case):}
\begin{equation*}\label{m1}
    y_i=\sin(2x_{i1})+2\exp(-16x_{i2}^2)+0.5e_i,
\end{equation*}
where $x_{i1}$ and $x_{i2}$ are independently generated from $N(0,0.5^2)$.
\item []\textbf{(2) Model 2 (Heteroscedastic Case):} 
\begin{equation*}\label{m2}
    y_i=1+\sin(x_{i1})+\exp\big({0.5x_{i1}^2-x_{i1}x_{i2}+0.2x_{i2}^2}\big)+|\frac{1+0.2(x_{i1}+x_{i2})}{5}|e_i.
\end{equation*}
For other censoring, the model are denoted as follows:
where $x_{i1}\sim Unif(-1,1)$ and $x_{i2} \sim N(0,1)$, respectively.
\end{itemize}

To illustrate the robustness of our proposed method, we consider two error distributions for each scenario: a normal and a heavy-tailed distribution. Specifically, for the normal case, the errors are generated as $e_i \sim N(0, 1)$; for the heavy-tailed case, they follow a Student’s $t$-distribution with 3 degrees of freedom, $t(3)$. We also consider two censoring rates, 25\% and 50\%. Detailed parameter settings are summarized in Table \ref{tab:setting}.


\begin{table}[htbp]
    \small
  \centering
  \setlength{\tabcolsep}{2.5pt}
  \caption{The distributions of censoring rates for different censoring types of the two simulation scenarios.}
    \begin{tabular}{cccccc}
    \toprule
    \multirow{2}[4]{*}{\textbf{Model}} & \multirow{2}[4]{*}{\textbf{Error term}} & \multirow{2}[4]{*}{\textbf{Censoring rate}} & \multicolumn{3}{c}{\textbf{Distribution of censoring sample}} \\
\cmidrule{4-6}          &       &       & Right & Left  & Interval \\
    \midrule
    \multirow{4}[4]{*}{\textbf{Model 1}} & \multirow{2}[2]{*}{$N(0,1)$} & 25\%  & $N(1.4,2^2)$& $N(0,2^2)$& $L\sim N(-0.5,2^2), R\sim N(0,2^2)$ \\
          &       & 50\%  & $N(0.6,2^2)$& $N(0.6,2^2)$ & $L\sim N(0,2^2), R\sim N(1.5,2^2)$ \\
\cmidrule{2-6}          & \multirow{2}[2]{*}{$t(3)$} & 25\%  & $N(1.5,2^2)$& $N(0,2^2$)& $L\sim N(-0.5,2^2), R\sim N(0,2^2)$ \\
          &       & 50\%  & $N(0.65,2^2)$& $N(0.6,2^2)$& $L\sim N(0,2^2), R\sim N(1.5,2^2)$ \\
    \midrule
    \multirow{4}[4]{*}{\textbf{Model 2}} & \multirow{2}[2]{*}{$N(0,1)$} & 25\%  & $\exp(4)$ & $\exp(2)$& $L\sim \exp(0.85), R \sim \exp(1.35)$\\
          &       & 50\%  & $\exp(3)$& $\exp(3)$& $L\sim \exp(0.55), R\sim \exp(1.45)$\\
\cmidrule{2-6}          & \multirow{2}[2]{*}{$t(3)$} & 25\%  & $\exp(4)$ & $\exp(2)$& $L\sim \exp(0.85), R \sim \exp(1.35)$\\
          &       & 50\%  & $\exp(3)$& $\exp(3)$& $L\sim \exp(0.55), R\sim \exp(1.45)$\\
    \bottomrule
    \end{tabular}
  \label{tab:setting}
\end{table}

In each simulation, we generate $n = 1000$ samples and randomly split the data into 80\% for training and 20\% for evaluating predictive performance. The primary evaluation metrics are the Expectile Loss (EL) and the Expectile Loss Ratio (EL$_{\text{ratio}}$), defined as:
\begin{equation*}
    \begin{aligned}
    \text{EL} = \frac{1}{n_\text{test}}\sum_{i=1}^{n_\text{test}}\rho_{\tau}\big(y_i-\hat{y}_i\big),  \ \ \   &\text{EL}_\text{ratio} = \frac{\text{EL}_\text{DAERNN}}{\text{EL}_\text{Compete}},
    \end{aligned}
\end{equation*}
where $n_\text{test}$ is the sample size of test set, $\text{EL}_\text{DAERNN}$ and $\text{EL}_\text{Compete}$ represent $\text{EL}$ values obtained through \textsf{DAERNN} and other three competitive methods (\textsf{FULL}, \textsf{DALinear} and \textsf{WERNN}). We report the average values of these evaluation metrics under each scenario, based on $200$ replications and across $9$ expectile levels, i.e., $\tau = \{0.1, 0.2, \ldots, 0.9\}$.

Additionally, our proposed method \textsf{DAERNN}, along with two other neural networks-based methods, requires hyperparameter tuning for model construction. The hyperparameters considered include the number of layers $L \in \{2, 3, 4\}$, number of nodes per layer $J \in \{16, 32, 64\}$, learning rate $\in \{0.01, 0.1\}$, dropout rate $\in \{0.1, 0.2, 0.3\}$, number of training epochs $\in \{50, 100\}$, and batch size $\in \{64, 128, 256\}$. The activation function is ReLU. Hyperparameters are tuned using $5$-fold cross-validation on the training set. To reduce computational cost, tuning for all three neural network methods is performed only on the uncensored samples.

\subsection{Performance results} \label{sub:3-2}

\subsubsection{Results for right-censoring case}\label{sub:right}

\begin{figure}
    \centering
    \includegraphics[width=1\linewidth]{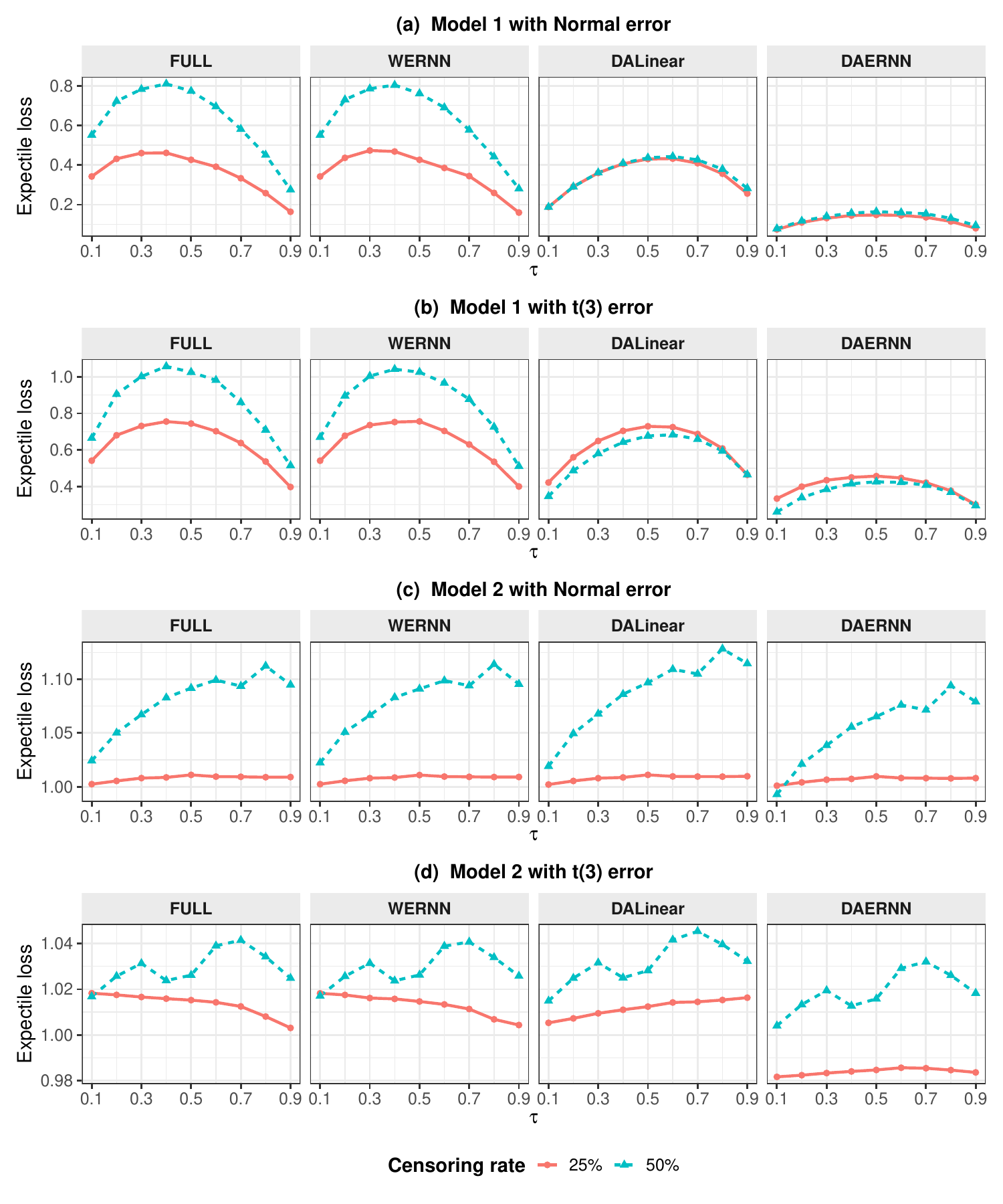}
    \caption{Expectile Losses of predicted responses for different methods and settings at $\tau = 0.1, 0.3, 0.5, 0.7,$ and $0.9$, under right censoring with different censoring rate.}
    \label{fig:EL}
\end{figure}

Table~\ref{tab:ER-r} reports the EL$_{\text{ratio}}$ values under both homoscedastic and heteroscedastic settings, across different error distributions and censoring rates of 25\% and 50\%. The EL$_{\text{ratio}}$ is defined as the ratio of the expectile loss obtained from \textsf{DAERNN} to that from the \textsf{FULL}, \textsf{DALinear}, and \textsf{WERNN} methods. Hence, a ratio below one indicates that \textsf{DAERNN} performs better than the corresponding competitive method, while a ratio above one suggests inferior performance. Figure~\ref{fig:EL} illustrates the expectile loss ($\text{EL}$) across different model settings and methods, under varying censoring rates.

Our main findings are summarized as follows:
\begin{itemize}
    \item[1)] Overall, \textsf{DAERNN} consistently outperforms all competitive methods in terms of predictive accuracy. Compared with \textsf{FULL} and \textsf{WERNN}, its predictive accuracy improves by approximately 50\% when the censoring rate is 25\%. This finding highlights the importance of accounting for censoring effects and demonstrates the robustness of the data augmentation approach, which does not rely on the assumption of a correctly specified survival function.

    \item[2)] The predictive accuracy also shows substantial improvement compared with \textsf{DAlinear}, with gains of at least 20\%. This demonstrates that replacing the linear model structure with a neural network can significantly enhance predictive performance, particularly for data with complex underlying relationships.
    
    \item[3)] Notably, the performance remains comparable even under heavy-tailed error distributions and heterogeneous settings, which demonstrating the robustness of the proposed method.
            
    \item[4)] The \textsf{DAERNN} method demonstrates stable performance in terms of EL values across different censoring rates, expect for Model 2 under a $t(3)$ error distribution (Figure~\ref{fig:EL}). When compared against competing methods, this observed stability suggests that the censoring rate does not significantly influence the performance of DAERNN, especially within homogeneous settings.
\end{itemize}

\begin{figure}
    \centering
    \includegraphics[width=1\linewidth]{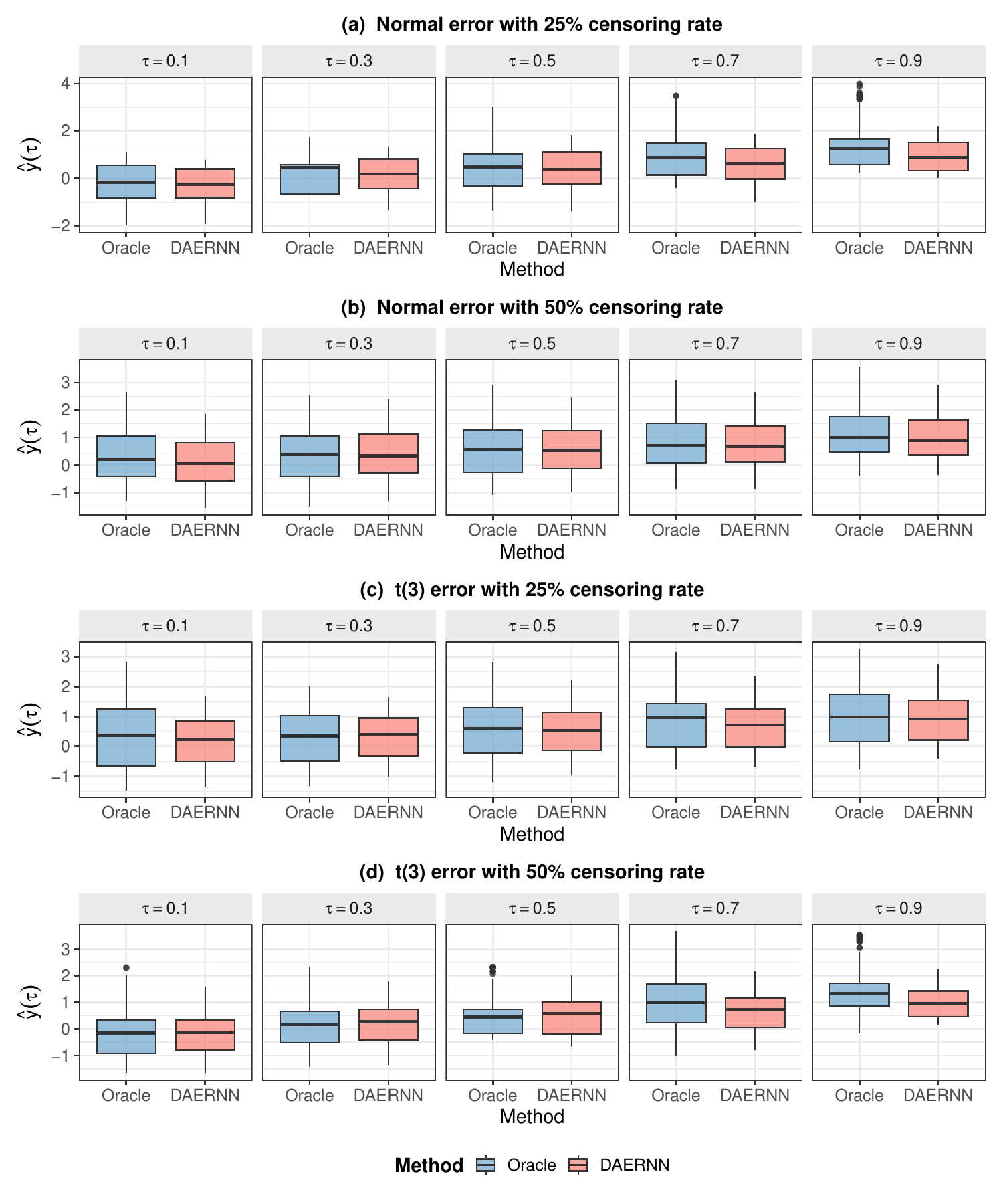}
    \caption{The boxplots of the predicted responses at $\tau = 0.1, 0.3, 0.5, 0.7, 0.9$ from the \textsf{Oracle} models and the $\textsf{DAERNN}$ under Model 1, across different error distributions and censoring rates.}
    \label{fig:ypred-M1}
\end{figure}

\begin{figure}
    \centering
    \includegraphics[width=1\linewidth]{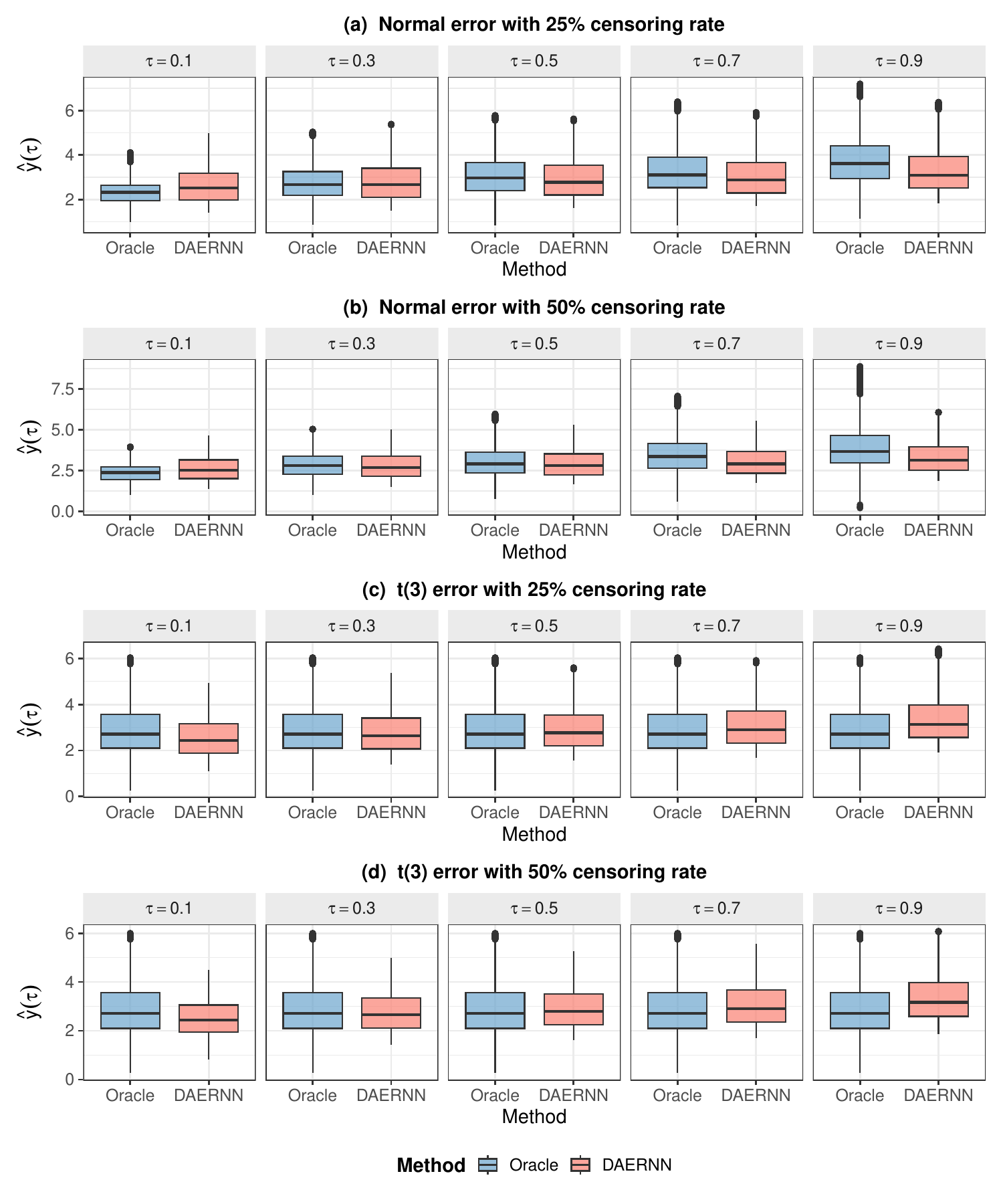}
    \caption{The boxplots of the predicted responses at $\tau = 0.1, 0.3, 0.5, 0.7, 0.9$ from the \textsf{Oracle} models and the $\textsf{DAERNN}$ under Model 2, across different error distributions and censoring rates.}
    \label{fig:ypred-M2}
\end{figure}

In addition, we also consider the results of the ERNNs model fitted without censoring in the dataset (referred to as \textsf{Oracle}) and compare its performance with our proposed method. Figure~\ref{fig:ypred-M1} and Figure~\ref{fig:ypred-M2} illustrate the distribution of the predicted values $\hat{y}(\tau)$ from the \textsf{Oracle} and \textsf{DAERNN} models under Model~1 and Model~2, respectively. The distributions of the predicted responses are quite similar, indicating that the performance of our proposed method can achieve comparable accuracy to that of the complete-case model.

\begin{table}[htbp]
  \centering
  \small
  \renewcommand{\arraystretch}{0.8}
   \caption{The $\text{EL}_{\text{ratio}}$ values for errors following $N(0,1)$ and $t(3)$ distribution, comparing the proposed \textsf{DAERNN} with the \textsf{FULL}, \textsf{WERNN}, and \textsf{DALinear} methods at $\tau=0.1,0.3,0.5,0.7,0.9$. The results are based on right-censoring with censoring rates of 25\% and 50\%.}
    \begin{tabular}{c|c|cccccccc}
    \toprule
    & & \multicolumn{4}{c}{\textbf{Model 1}} & \multicolumn{4}{c}{\textbf{Model 2}} \\
    \cmidrule(lr){3-6} \cmidrule(lr){7-10}
    \multirow{-1}{*}{\textbf{Method}} & \multirow{-1}{*}{$\boldsymbol{\tau}$}
                   & \multicolumn{2}{c}{$N(0,1)$} & \multicolumn{2}{c}{$t(3)$}
                   & \multicolumn{2}{c}{$N(0,1)$} & \multicolumn{2}{c}{$t(3)$} \\
    \cmidrule(lr){3-4} \cmidrule(lr){5-6} \cmidrule(lr){7-8} \cmidrule(lr){9-10}
                   & & {25\%} & {50\%} & {25\%} & {50\%}
                   & {25\%} & {50\%} & {25\%} & {50\%} \\
    \midrule
    \multirow{5}[2]{*}{\textsf{FULL/DAERNN}} & 0.1   & 0.223  & 0.145  & 0.544  & 0.378  & 0.667  & 0.617  & 0.686  & 0.640  \\
          & 0.3   & 0.292  & 0.182  & 0.555  & 0.377  & 0.690  & 0.641  & 0.699  & 0.659  \\
          & 0.5   & 0.360  & 0.218  & 0.589  & 0.410  & 0.712  & 0.665  & 0.719  & 0.686  \\
          & 0.7   & 0.430  & 0.272  & 0.642  & 0.468  & 0.746  & 0.698  & 0.741  & 0.711  \\
          & 0.9   & 0.532  & 0.368  & 0.742  & 0.569  & 0.795  & 0.758  & 0.792  & 0.778  \\
    \midrule
    \multirow{5}[2]{*}{\textsf{WERNN/DAERNN}} & 0.1   & 0.223  & 0.145  & 0.543  & 0.377  & 0.668  & 0.619  & 0.687  & 0.640  \\
          & 0.3   & 0.286  & 0.181  & 0.551  & 0.376  & 0.692  & 0.639  & 0.700  & 0.658  \\
          & 0.5   & 0.359  & 0.220  & 0.581  & 0.410  & 0.711  & 0.666  & 0.720  & 0.684  \\
          & 0.7   & 0.415  & 0.274  & 0.654  & 0.460  & 0.742  & 0.696  & 0.747  & 0.713  \\
          & 0.9   & 0.540  & 0.371  & 0.738  & 0.571  & 0.792  & 0.761  & 0.790  & 0.773  \\
    \midrule
    \multirow{5}[2]{*}{\textsf{DALinear/DAERNN}} & 0.1   & 0.400  & 0.424  & 0.729  & 0.738  & 0.776  & 0.687  & 0.793  & 0.711  \\
          & 0.3   & 0.367  & 0.391  & 0.628  & 0.651  & 0.759  & 0.661  & 0.765  & 0.681  \\
          & 0.5   & 0.346  & 0.378  & 0.595  & 0.619  & 0.744  & 0.649  & 0.755  & 0.672  \\
          & 0.7   & 0.334  & 0.362  & 0.585  & 0.606  & 0.733  & 0.633  & 0.738  & 0.651  \\
          & 0.9   & 0.317  & 0.336  & 0.615  & 0.611  & 0.714  & 0.615  & 0.715  & 0.630  \\
    \bottomrule
    \end{tabular}%
  \label{tab:ER-r}%
\end{table}%

\subsubsection{Results for other censoring types}\label{sub:other}
Table~\ref{tab:ER-l} reports the EL$_\text{ratio}$ values comparing \textsf{DAERNN} with both \textsf{FULL} and \textsf{DALinear} for left censoring. The EL$_\text{ratio}$ remains lower than 1 in most caseswhen compared to both \textsf{FULL} and \textsf{DALinear}, confirming the advantage of \textsf{DAERNN}. Moreover, the results show only minor variations across different error distributions at the same expectile level, indicating the robustness of the proposed method. Results for the interval censoring case are shown in Table~\ref{tab:ER-int}, which are consistent with those from the left censoring case; thus, we omit redundant discussion.

\begin{table}[htbp]
  \centering
  \small
  \renewcommand{\arraystretch}{0.8}
   \caption{The $\text{EL}_{\text{ratio}}$ values for errors following $N(0,1)$ and $t(3)$ distribution, comparing the proposed \textsf{DAERNN} with the \textsf{FULL}, \textsf{DALinear} methods at $\tau=0.1,0.3,0.5,0.7,0.9$. The results are based on left-censoring with censoring rates of 25\% and 50\%.}
    \begin{tabular}{c|c|cccccccc}
    \toprule
    & & \multicolumn{4}{c}{\textbf{Model 1}} & \multicolumn{4}{c}{\textbf{Model 2}} \\
    \cmidrule(lr){3-6} \cmidrule(lr){7-10}
    \multirow{-1}{*}{\textbf{Method}} & \multirow{-1}{*}{$\boldsymbol{\tau}$}
                   & \multicolumn{2}{c}{$N(0,1)$} & \multicolumn{2}{c}{$t(3)$}
                   & \multicolumn{2}{c}{$N(0,1)$} & \multicolumn{2}{c}{$t(3)$} \\
    \cmidrule(lr){3-4} \cmidrule(lr){5-6} \cmidrule(lr){7-8} \cmidrule(lr){9-10}
                   & & {25\%} & {50\%} & {25\%} & {50\%}
                   & {25\%} & {50\%} & {25\%} & {50\%} \\
    \midrule
    \multirow{5}[2]{*}{\textsf{FULL/DAERNN}} & 0.1   & 0.580  & 0.387  & 0.740  & 0.610  & 0.988  & 0.781  & 0.943  & 1.050  \\
          & 0.3   & 0.463  & 0.286  & 0.669  & 0.494  & 0.943  & 0.944  & 0.875  & 0.917  \\
          & 0.5   & 0.376  & 0.238  & 0.618  & 0.435  & 0.803  & 0.613  & 0.970  & 0.722  \\
          & 0.7   & 0.309  & 0.191  & 0.573  & 0.401  & 0.740  & 0.607  & 0.885  & 0.583  \\
          & 0.9   & 0.229  & 0.148  & 0.548  & 0.398  & 0.895  & 0.740  & 0.995  & 0.624  \\
    \midrule
    \multirow{5}[2]{*}{\textsf{DALinear/DAERNN}} & 0.1   & 0.387  & 0.402  & 0.668  & 0.687  & 0.846  & 0.311  & 0.485  & 0.324  \\
          & 0.3   & 0.356  & 0.371  & 0.595  & 0.627  & 0.547  & 0.203  & 0.425  & 0.259  \\
          & 0.5   & 0.337  & 0.358  & 0.581  & 0.614  & 0.291  & 0.171  & 0.254  & 0.163  \\
          & 0.7   & 0.326  & 0.345  & 0.581  & 0.616  & 0.181  & 0.127  & 0.174  & 0.128  \\
          & 0.9   & 0.320  & 0.338  & 0.661  & 0.680  & 0.099  & 0.100  & 0.119  & 0.091  \\
    \bottomrule
    \end{tabular}%
  \label{tab:ER-l}%
\end{table}%

\begin{table}[htbp]
   \centering
  \small
  \renewcommand{\arraystretch}{0.8}
   \caption{The $\text{EL}_{\text{ratio}}$ values for errors following $N(0,1)$ and $t(3)$ distribution, comparing the proposed \textsf{DAERNN} with the \textsf{FULL}, \textsf{DALinear} methods at $\tau=0.1,0.3,0.5,0.7,0.9$. The results are based on interval-censoring with censoring rates of 25\% and 50\%.}
    \begin{tabular}{c|c|cccccccc}
    \toprule
    & & \multicolumn{4}{c}{\textbf{Model 1}} & \multicolumn{4}{c}{\textbf{Model 2}} \\
    \cmidrule(lr){3-6} \cmidrule(lr){7-10}
    \multirow{-1}{*}{\textbf{Method}} & \multirow{-1}{*}{$\boldsymbol{\tau}$}
                   & \multicolumn{2}{c}{$N(0,1)$} & \multicolumn{2}{c}{$t(3)$}
                   & \multicolumn{2}{c}{$N(0,1)$} & \multicolumn{2}{c}{$t(3)$} \\
    \cmidrule(lr){3-4} \cmidrule(lr){5-6} \cmidrule(lr){7-8} \cmidrule(lr){9-10}
                   & & {25\%} & {50\%} & {25\%} & {50\%}
                   & {25\%} & {50\%} & {25\%} & {50\%} \\
    \midrule
    \multirow{5}[2]{*}{\textsf{FULL/DAERNN}} & 0.1   & 0.638  & 0.499  & 0.890  & 0.810  & 0.974  & 1.122  & 0.943  & 1.194  \\
          & 0.3   & 0.776  & 0.625  & 0.903  & 0.845  & 0.973  & 0.641  & 0.875  & 0.927  \\
          & 0.5   & 0.821  & 0.711  & 0.921  & 0.878  & 0.996  & 0.877  & 0.970  & 0.924  \\
          & 0.7   & 0.858  & 0.747  & 0.933  & 0.898  & 0.954  & 0.985  & 0.885  & 0.939  \\
          & 0.9   & 0.851  & 0.717  & 0.941  & 0.898  & 1.008  & 1.159  & 0.995  & 0.932  \\
    \midrule
    \multirow{5}[2]{*}{\textsf{DALinear/DAERNN}} & 0.1   & 0.391  & 0.392  & 0.708  & 0.709  & 0.722  & 1.243  & 0.639  & 0.918  \\
          & 0.3   & 0.350  & 0.354  & 0.614  & 0.624  & 0.668  & 0.750  & 0.594  & 0.684  \\
          & 0.5   & 0.333  & 0.344  & 0.579  & 0.593  & 0.636  & 0.690  & 0.515  & 0.612  \\
          & 0.7   & 0.320  & 0.334  & 0.579  & 0.595  & 0.620  & 0.678  & 0.520  & 0.597  \\
          & 0.9   & 0.315  & 0.328  & 0.643  & 0.657  & 0.600  & 0.585  & 0.453  & 0.551  \\
    \bottomrule
    \end{tabular}%
  \label{tab:ER-int}%
\end{table}%

\subsection{Computational cost}
Here we compare the computation time of different method methods under different scenarios.
\begin{table}[htbp]
  \centering
  \small
  \renewcommand{\arraystretch}{0.8}
  \caption{Average computational time (secs)}
    \begin{tabular}{cccccccc}
    \toprule
    \multirow{2}{*}{\textbf{Error term}} & \multirow{2}{*}{\textbf{Censoring rate}} 
    & \multicolumn{3}{c}{\textbf{Model 1}} 
    & \multicolumn{3}{c}{\textbf{Model 2}} \\
    \cmidrule(lr){3-5}\cmidrule(lr){6-8}
    & & \textsf{FULL}  & \textsf{DAERNN} & \textsf{WERNN} & \textsf{FULL}  & \textsf{DAERNN} & \textsf{WERNN} \\
    \midrule
    \multirow{2}[2]{*}{$N(0,1)$} & 25\%  & 7.240  & 23.426  & 132.676  & 5.698  & 20.471  & 147.545  \\
          & 50\%  & 6.641  & 21.368  & 131.773  & 5.208  & 17.687  & 150.272  \\
    \midrule
    \multirow{2}[2]{*}{$t(3)$} & 25\%  & 6.341  & 18.164  & 124.364  & 6.198  & 19.295  & 129.840  \\
          & 50\%  & 6.198  & 19.295  & 129.840  & 5.522  & 18.779  & 156.787  \\
    \bottomrule
    \end{tabular}%
  \label{tab:time}%
\end{table}%

Table~\ref{tab:time} reports the average computational time of \textsf{DAERNN} and other competing methods. The results for \textsf{DALinear} are omitted, as it operates under a linear framework and is not directly comparable with the neural networks-based approaches. As observed, the proposed \textsf{DAERNN} requires slightly more time than the \textsf{FULL} model but is considerably more efficient than \textsf{WERNN}. This is reasonable since \textsf{DAERNN} involves iterative imputation of censored observations, which naturally increases computational cost. Nevertheless, the results indicate that \textsf{DAERNN} achieves higher accuracy than \textsf{FULL} with only a modest increase in computation time, and it significantly outperforms \textsf{WERNN} in both accuracy and efficiency, highlighting the advantages of data-augmentation-based strategies.

The optimal hyperparameters number of layers ($L$) and number of nodes per layer ($J$) for expectile levels $\tau \in \{0.3, 0.5, 0.7\}$ selected via 5-fold cross-validation, are reported in Table~\ref{tab:hyper}. The results suggest that the proposed method performs well with a relatively simple neural network structure, indicating that \textsf{DAERNN} is practical and efficient for real applications.
\begin{table}[htbp]
  \centering
  \small
  \renewcommand{\arraystretch}{0.78}
  \caption{The selected hyperparameters: number of layers ($L$) and number of nodes per layer ($J$) for \textsf{DAERNN}, obtained via 5-fold cross validation on the simulation data.}
    \begin{tabular}{ccccccccccccccc}
   \toprule
\multirow{3}{*}{\textbf{Error term}} & \multirow{3}{*}{\textbf{Censoring rate}} & \multirow{3}{*}{\boldsymbol{$\tau$}} & \multicolumn{6}{c}{\textbf{Model 1}} & \multicolumn{6}{c}{\textbf{Model 2}} \\
\cmidrule(lr){4-9} \cmidrule(lr){10-15} 
& & & \multicolumn{2}{c}{right} & \multicolumn{2}{c}{left} & \multicolumn{2}{c}{interval} & \multicolumn{2}{c}{right} & \multicolumn{2}{c}{left} & \multicolumn{2}{c}{interval} \\
\cmidrule(lr){4-5} \cmidrule(lr){6-7} \cmidrule(lr){8-9} \cmidrule(lr){10-11} \cmidrule(lr){12-13} \cmidrule(lr){14-15}
& & & $L$ & $J$ & $L$ & $J$ & $L$ & $J$ & $L$ & $J$ & $L$ & $J$ & $L$ & $J$ \\
\midrule
    \multirow{6}[4]{*}{$N(0,1)$} & \multirow{3}[2]{*}{25\%} & 0.3   & 3     & 32    & 3     & 16    & 2     & 32    & 4     & 16    & 2     & 16    & 3     & 32 \\
          &       & 0.5   & 4     & 16    & 4     & 16    & 3     & 32    & 3     & 32    & 3     & 32    & 3     & 32 \\
          &       & 0.7   & 3     & 32    & 2     & 16    & 4     & 32    & 4     & 64    & 2     & 32    & 4     & 32 \\
\cmidrule{2-15}          & \multirow{3}[2]{*}{50\%} & 0.3   & 3     & 16    & 3     & 32    & 2     & 32    & 4     & 64    & 2     & 32    & 2     & 32 \\
          &       & 0.5   & 3     & 16    & 3     & 32    & 4     & 16    & 3     & 32    & 4     & 32    & 2     & 32 \\
          &       & 0.7   & 3     & 32    & 3     & 32    & 4     & 16    & 4     & 32    & 4     & 64    & 4     & 16 \\
    \midrule
    \multirow{6}[4]{*}{$t(3)$} & \multirow{3}[2]{*}{25\%} & 0.3   & 3     & 32    & 2     & 16    & 2     & 32    & 4     & 32    & 3     & 32    & 2     & 32 \\
          &       & 0.5   & 4     & 16    & 4     & 32    & 3     & 16    & 3     & 32    & 4     & 32    & 3     & 16 \\
          &       & 0.7   & 2     & 16    & 2     & 16    & 2     & 32    & 3     & 16    & 4     & 32    & 4     & 16 \\
\cmidrule{2-15}          & \multirow{3}[2]{*}{50\%} & 0.3   & 4     & 32    & 3     & 16    & 3     & 32    & 4     & 16    & 2     & 32    & 3     & 32 \\
          &       & 0.5   & 3     & 32    & 2     & 64    & 4     & 16    & 3     & 64    & 4     & 32    & 4     & 32 \\
          &       & 0.7   & 2     & 16    & 4     & 16    & 3     & 16    & 4     & 32    & 4     & 64    & 4     & 32 \\
    \bottomrule
    \end{tabular}%
  \label{tab:hyper}%
\end{table}%

\section{Empirical study}
\label{sec:realdata}
For illustration purpose, we apply the proposed method to analyze data sets from two practical examples.

\subsection{WHAS dataset}
The WHAS dataset was collected from the Worcester Heart Attack Study, conducted between 1975 and 2001. This study investigates the effects of various factors on the survival time following hospital admission for acute myocardial infarction. The dataset contains 500 observations and 14 covariates, including gender, age, and congestive heart complications, among others. The censoring rate is approximately 57.0\%. The data are available in the R package \texttt{smoothHR} and also used the in outer study \citep{zhang2024weighted}. Description of variables are shown in Table \ref{tab:whas-var}.
\begin{table}[htbp]
    \centering
    \small
  \renewcommand{\arraystretch}{0.78}
  \caption{Description of variables for WHAS dataset.}
\begin{tabular}{ccl} 
\toprule
\textbf{Variable} & \textbf{Description}  &\textbf{Variable declaration}\\ 
\midrule
\textit{Survival time} & Length of follow-up days  &Depend variable\\
\textit{Status} & Status as of last follow-up&1 = Dead, 0 = Alive\\
\textit{Gender} & Gender&0 = Male, 1 = Female\\
\textit{cvd} & History of cardiovascular disease&0 = No, 1 = Yes\\
\textit{afb} & Atrial fibrillation&0 = No, 1 = Yes\\
\textit{sho} & Cardiogenic shock&0 = No, 1 = Yes\\
\textit{chf} & Congestive heart complications&0 = No, 1 = Yes\\
\textit{av3} & Complete heart block&0 = No, 1 = Yes\\
\textit{miord} & MI Order&0 = First, 1 = Recurrent\\
\textit{mitype} & MI Type&0 = non Q-wave, 1 = Q-wave\\
\textit{Age} & Age at hospital admission (years)  &\\
\textit{hr} & Initial heart rate. Beats per minute  &\\
\textit{los} & Length of hospital stay  &\\
\textit{sysbp} & Initial systolic blood pressure (mmHg)  &\\
\textit{diasbp} & Initial diastolic blood pressure (mmHg)  &\\
\textit{bmi} & Body mass index  &\\
\bottomrule
\end{tabular}
  \label{tab:whas-var}%
\end{table}

In this section, we compare the predictive performance of the proposed method with \textsf{WERNN} and \textsf{DALinear} using 10-fold cross-validation, computing the EL$_\text{ratio}$ for each method across $\tau\in\{0.1,0.2,\cdots,0.9\}$. The empirical results are summarized in Table~\ref{tab:whas}. As shown, \textsf{DAERNN} generally achieves lower forecasting errors, further confirming the superior predictive capability of the proposed approach.

\begin{table}[htbp]
\small
  \renewcommand{\arraystretch}{0.8}
  \centering
  \caption{Expectile Loss Ratio for WHAS data by 10-fold cross-validation.}
    \begin{tabular}{cccccccccc}
    \toprule
    $\boldsymbol{\tau}$   & 0.1   & 0.2   & 0.3   & 0.4   & 0.5   & 0.6   & 0.7   & 0.8   & 0.9 \\
    \midrule
    \textsf{DALinear} & 0.717  & 0.760  & 0.828  & 0.856  & 0.911  & 0.895  & 0.961  & 0.981  & 1.035  \\
    \textsf{WERNN} & 1.024  & 0.875  & 0.848  & 0.878  & 0.865  & 0.875  & 0.849  & 0.840  & 0.874  \\
    \bottomrule
    \end{tabular}%
  \label{tab:whas}%
\end{table}%

\subsection{YVR dataset}
To evaluate the performance of the proposed method under complex censoring mechanisms, we apply the \textsf{DAERNN} approach to a real-world dataset and artificially generate censored samples at different censoring rates. The data, available in the R package \texttt{qrnn} as the \texttt{“YVRprecip”} object, contain daily precipitation totals (in mm) recorded at Vancouver International Airport (YVR) from 1971 to 2000. The dataset consists of $n = 10{,}958$ observations and includes three covariates: daily time series of sea-level pressure, 700-hPa specific humidity, and 500-hPa geopotential height. To account for seasonal effects relevant to precipitation downscaling, we also include sine and cosine transformations of the day of the year as additional predictors. Similarly, here we make the same comparisons as those in the simulation studies.


The ``YVRprecip’’ dataset is originally uncensored, we follow the instruction of \citet{hao2022data} artificially introduce right, left, and interval censoring to assess model performance under different censoring scenarios. The specific settings for each censoring type and rate are summarized in Table~\ref{tab:YVR}.

\begin{table}[htbp]
\small
  \renewcommand{\arraystretch}{0.8}
  \setlength{\tabcolsep}{4.5pt} 
  \centering
  \caption{Expectile Loss Ratio for right-censoring case of YVR dataset by 10-fold cross-validation.}
    \begin{tabular}{c|c|ccccccccc}
   \toprule
\multirow{2}[3]{*}{\textbf{Method}} 
& \multicolumn{1}{c|}{\multirow{2}[3]{*}{\shortstack{\textbf{Censoring}\\\textbf{rate}}}} 
& \multicolumn{9}{c}{$\boldsymbol{\tau}$} \\ 
\cmidrule{3-11}
& & 0.1 & 0.2 & 0.3 & 0.4 & 0.5 & 0.6 & 0.7 & 0.8 & 0.9 \\
\midrule
    \multirow{2}[0]{*}{\textsf{Oracle/DAERNN}} & 25\%  & 1.050  & 1.106  & 1.177  & 1.212  & 1.278  & 1.342  & 1.455  & 1.566  & 1.995  \\
          & 50\%  & 1.088  & 1.132  & 1.176  & 1.276  & 1.379  & 1.611  & 1.821  & 2.064  & 2.732  \\
    \midrule
    \multirow{2}[1]{*}{\textsf{OWERNN/DAERNN}} & 25\%  & 0.985  & 1.001  & 1.007  & 0.975  & 0.967  & 0.984  & 1.001  & 1.003  & 0.968  \\
          & 50\%  & 0.722  & 0.761  & 0.788  & 0.802  & 0.820  & 0.839  & 0.853  & 0.858  & 0.897  \\
    \midrule
    \multirow{2}[1]{*}{\textsf{ODALinear/DAERNN}} & 25\%  & 0.927  & 0.897  & 0.873  & 0.850  & 0.833  & 0.818  & 0.809  & 0.793  & 0.775  \\
          & 50\%  & 0.943  & 0.923  & 0.911  & 0.902  & 0.895  & 0.889  & 0.883  & 0.870  & 0.859  \\
    \bottomrule
    \end{tabular}%
  \label{tab:YVR}%
\end{table}%

We evaluate and compare the values EL$_\text{ratio}$ using a 10-fold cross-validation to assess the relative performance of different methods and the results are shown in Table~\ref{tab:YVR}. The results show that the proposed method performs comparably to the \textsf{Oracle}, with $\text{EL}_{\text{ratio}}$ values close to 1 in most cases. Compared to the other two censored expectile methods, \textsf{DAERNN} demonstrates better performance, especially when the censoring rate is high.

\begin{figure}[ht]
    \centering
    \includegraphics[width=1\linewidth]{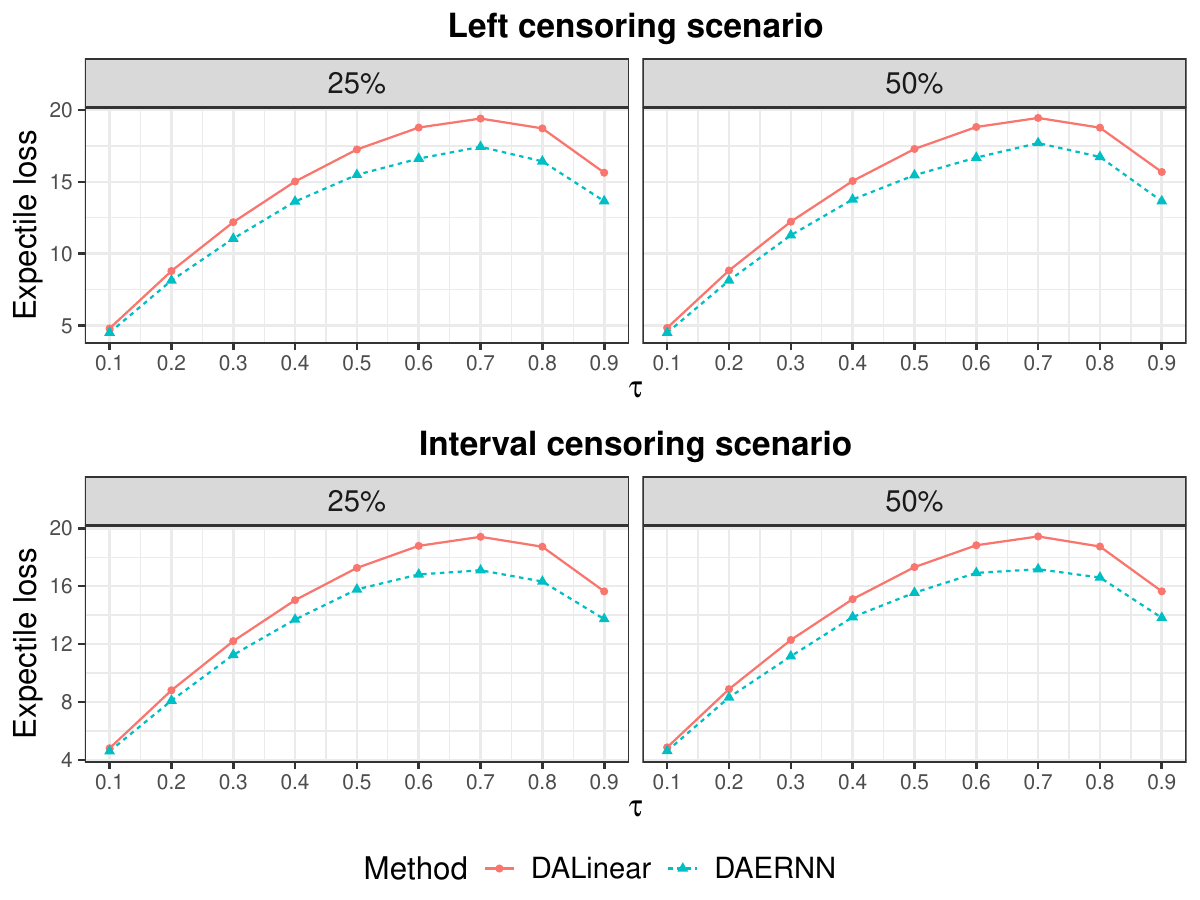}
     \caption{The expectile loss ($\text{EL}$) of \textsf{Oracle},\textsf{DALinear} and \textsf{DAERNN} method for left-censoring case top panel and interval- censoring scenario bottom panle, with 25\% and 50\% censor rates.}
     \label{fig:real-EL}
\end{figure}

Figure~\ref{fig:real-EL} demonstrates the expectile loss for other censoring scenarios. The result show that \textsf{DAERNN} is superior than \textsf{DALinear}. All in all, the precipitation dataset demonstrates the high prediction efficiency of our proposed \textsf{DAERNN} method.

\section{Conclusion and discussions}
\label{sec:discussion}

In this paper, we propose a unified estimation framework, \textsf{DAERNN}, for censored expectile regression neural network models under various censoring types. By integrating data augmentation techniques, \textsf{DAERNN} imputes censored outcomes while effectively capturing complex nonlinear and heterogeneous structures through expectile regression with neural networks. Both simulation and empirical studies demonstrate that \textsf{DAERNN} achieves performance comparable to the oracle model and significantly outperforms existing methods, including IPW-based ERNNs and linear data augmentation approaches.

Despite its advantages, several limitations merit future attention. First, like many neural network models, \textsf{DAERNN} lacks interpretability. Incorporating semiparametric structures, such as single-index models, partial linear model may enhance model transparency while preserving flexibility \citep{zhong2021deep}. Second, although neural networks can accommodate high-dimensional data through feature extraction in hidden layers, their performance is constrained by limited sample sizes. Future work may integrate variable screening or marginal feature selection to improve model efficiency in high-dimensional settings.

\section*{Acknowledgments}
This research was financially supported by the State Key Program of National Natural Science Foundation of China [Nos. 72531002].

\bibliography{mybib}
\bibliographystyle{agsm}

\end{document}